\documentclass[sigconf,nonacm]{acmart}
\usepackage{graphicx}
\usepackage{subfigure}
\usepackage{makecell}
\usepackage{amsmath}
\usepackage{amsfonts}
\usepackage{algorithm}
\usepackage{setspace}
\usepackage{dblfloatfix} 
\usepackage{mathrsfs}
\usepackage{algorithmic}

\usepackage{balance}
\usepackage{xcolor}
\theoremstyle{definition}
\newtheorem{definition}{Definition}
\newcommand{\hup}{\vspace*{-0.1cm}}

\AtBeginDocument{%
  \providecommand\BibTeX{{%
    \normalfont B\kern-0.5em{\scshape i\kern-0.25em b}\kern-0.8em\TeX}}}
\copyrightyear{2022}
\acmYear{2022}
\setcopyright{acmlicensed}\acmConference[CIKM '22]{31st ACM International Conference on Information and Knowledge Management}{October 17--22, 2022}{Atlanta, Georgia, the USA}
\acmBooktitle{31st ACM International Conference on Information and Knowledge Management (CIKM '22), October 17--22, 2022, Atlanta, Georgia, the USA}

\usepackage[T1]{fontenc}
\usepackage[utf8]{inputenc}
\author{Fang He}
\affiliation{\institution{The Pennsylvania State University}}
\email{fxh35@cse.psu.edu}

\author{Tao-Yang Fu}
\affiliation{\institution{The Pennsylvania State University}}
\email{txf225@psu.edu}

\author{Wang-chien Lee}
\affiliation{\institution{The Pennsylvania State University}}
\email{wlee@cse.psu.edu}

\setcopyright{none}             
\renewcommand\footnotetextcopyrightpermission[1]{} 

\begin{document}
\title{TraveL: Transformer-based Multi-view Path Distributional Representation Learning
}

\begin{abstract}

Path representation learning (PRL) for road networks has received increasing research attention, due to various path-related applications.
Existing works on PRL typically exploit the co-occurrence relationship among road segments and paths to learn a vector as the path representation, without exploring the varied traveler behaviors and the regional correlation on the path.
In this work, we propose to learn \textit{distributional representations}, which provide valuable information for use in
path-related applications, by capturing the \textit{varied traveler behaviors} as well as the \textit{various dependencies within regions of road segments}.
We propose a novel Transformer-based Multi-view Distributional Representation Learning (TraveL) framework to encode a path along with a travel starting time to a distributional representation, which can be used to decode possible samples of on-path traveler behavior.
Moreover, by analyzing the \textit{regional correlation} which reveals various road segment relationships, we propose a \textit{regional attention} to encode these correlations in a path.
Also, we explore the idea of Kolmogorov–Smirnov (K-S) test to compare the sampled traveler behavior against the collected ground truth to facilitate training.
Experimental results show that the proposed TraveL model outperforms the state-of-the-art methods on both synthetic and real-world datasets, by 14.7\% in Mean K-S distance for travel time distribution estimation, 16.7\% in Mean Absolute Error (MAE) for path similarity prediction, and 3.97\% in MAE for destination prediction.
\end{abstract}

\keywords{Path representation learning; Distributional representation; Regional attention}

\maketitle

\pagestyle{plain}               

\section{Introduction}
With the rapid growth of GPS-enabled devices and location-aware applications, a large volume of trajectory data are collected, which provides opportunities to study and improve various applications in intelligent transportation systems (ITS). 
Among them, many ITS applications involve \textit{paths}, i.e., a sequence of consecutive road segments, such as path travel time estimation, path recommendation, and destination prediction (given a partial path).\footnote{Note that a path is \textit{not} a trajectory which is a sequence of GPS points.} 
In support of these applications, in this work, we study the problem of \textit{Path Representation Learning (PRL)}, aiming to encode a given path in the road network into a general-purpose representation, e.g., a low-dimension latent vector. 
Similar to existing representation learning models in different domains, e.g., Doc2Vec for document representation learning \cite{le2014distributed} and HIN2Vec for network representation learning \cite{fu2017hin2vec}, PRL may reduce both storage cost and human labors required for extensive feature engineering in path-related applications, showing its great values.

\begin{figure}[t]
\hup\hup
\centering
\includegraphics[width=2.4in]{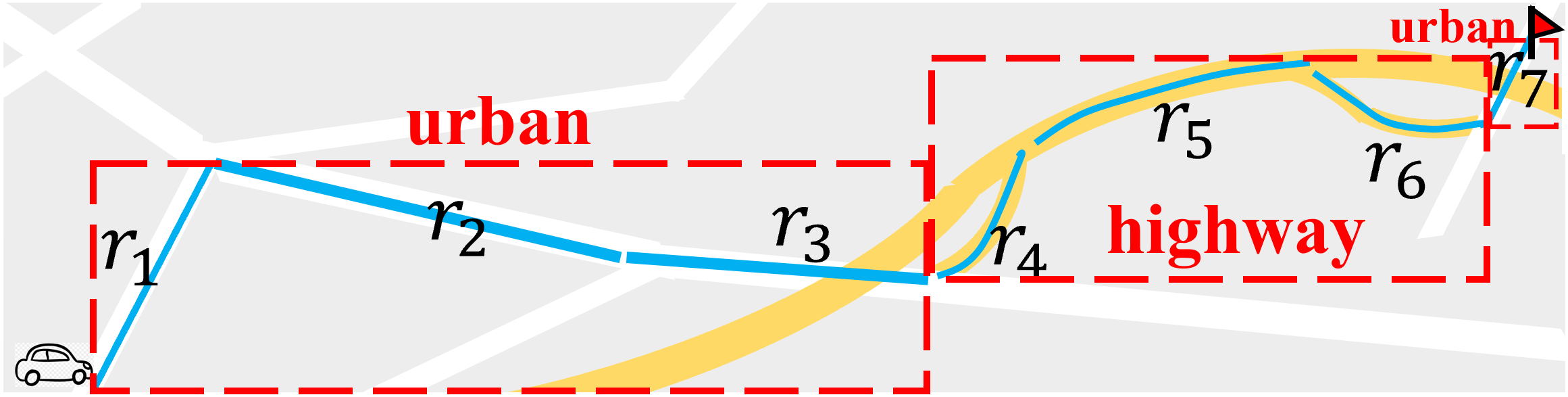}
\vspace{-0.12in}
\caption{Regional Correlation of a Path in Highway View}\centering
\vspace{-0.26in}
\label{fig:path-highway}
\end{figure}

For PRL in a road network, two kinds of information are important and essential: i) \textit{the varied behaviors from travelers} on the same path.
In addition to the static features of a path (e.g., the road network structure), it's imperative to capture the varied traveler behaviors, especially in terms of travel speed along the path, which intuitively benefits the travel time distribution estimation. 
It may also benefit path similarity estimation, especially when the similarity measurement of interest is related to the traveler behaviors on the path (e.g., finding another path with a similar travel time fluctuation as a given path);
and ii) \textit{regional correlation in the paths}, brought by the co-occurrence and same-type relationships among road segments, indicating various dependencies among the travel behaviors on different road segments.
For example, as shown in Figure~\ref{fig:path-highway}, drivers drive along a path consisting of 7 road segments (denoted as $r_1$ to $r_7$ respectively). We observe that $r_1$, $r_2$ and $r_3$ are in the urban area, while $r_4$, $r_5$ and $r_6$ are on the highway. We argue that the driving behavior, e.g., speed, from an urban road to a highway may change significantly, while the driving behavior on continuous road segments of the same type (i.e., a path region of highway road segments) is relatively stable. It would be helpful, yet challenging, to capture the regional correlation in paths for PRL.

Existing works on PRL starts by learning task-specific path representations. Liu et al.~\cite{liu2017semantic}, Li et al. ~\cite{li2017deepcas} and Yang et al. ~\cite{yang2019pathrank}
exploit various recurrent neural network based models to aggregate road segment embeddings in the path as the path representation for their targeted tasks.
Aiming for general-purpose PRL, Yang et al. propose Path InfoMax (PIM)~\cite{yang2021unsupervised}, with a curriculum negative sampling strategy to generate negative paths to learn path representations in a general graph.
Trembr, recently proposed by Fu et al.~\cite{fu2020trembr}, exploits pre-trained road segment embeddings to capture road segment co-occurrence and same-type relationship, and encodes a path into its representation with an LSTM-based encoder-decoder structure. 
PIM and Trembr do not capture the varied traveler behaviors, and thus fail to learn path representations well.

In this work, we propose to capture both the varied traveler behaviors and the regional correlation for PRL. 
We propose to explore an encoder-decoder framework which first encodes the targeted path into a probabilistic distribution as the representation, capable of generating expected travel behaviors on the path. 
However, we face \textit{three issues}: i) conventional latent vectors are insufficient to cover the varied traveler behaviors.
To address the issue, we explore the idea of \textit{distributional representation}, which learns a distribution as the representation of a path, with higher capacity than a latent vector. ii) the decoded travel speeds/behaviors of different travelers along a path do not follow a known parameterized distribution, making it inappropriate to learn the parameters based on some known distribution of travel behaviors (in speed). 
Instead, we propose to generate a set of \textit{on-path sequences (OP-Seqs)} from the distributional representation, each of which refers to a possible travel trace on the path, i.e., a trace consists of the path and the travel time spent on each road segment in the path. 
As such, we can optimize the path distributional representation by measuring and minimizing the dissimilarity between the generated and ground-truth OP-Seqs. iii) to bridge from the distributional representation to the OP-Seqs is challenging. To resolve the issue, we explore a sampling-based approach in the decoding phase, which regards a point sampled from the distributional representation as the representation of a possible OP-Seq. 
As such, the distributional representation covers all possible OP-Seqs and thus acts as a proper path representation.




In addition, we capture various correlations in path regions for PRL where a path region is some sequence of consecutive road segments in the path.  
We propose two schemes to form path regions to explore regional correlation: i) road-type based. The traveler behaviors in a region of a same road type (shown in Figure~\ref{fig:path-highway} with red bounding boxes) may be correlated; and ii) hop based, traveler behaviors within a number of hops along the path, may be correlated.
We use the term, \textit{view}, to denote the path structure exhibiting a specific regional correlation, e.g., highway view as shown in Figure~\ref{fig:path-highway}.
To capture the regional correlation, we propose the idea of \textit{regional attention}, by
letting each road segment correlate other road segments in the same region to enhance the road segment representations.


\begin{figure}[t]
\centering
\includegraphics[width=3.2in]{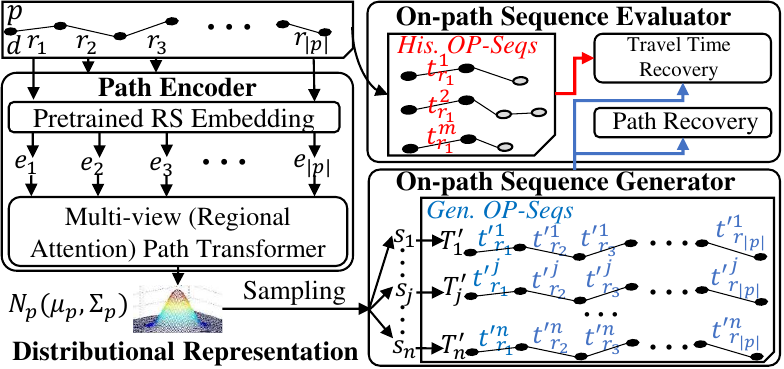}
\vspace{-0.15in}
\caption{The TraveL Framework}\centering
\label{fig:TRAVEL-Framework}
\vspace{-0.35in}
\end{figure}

To realize our ideas, we propose the \underline{Tra}nsformer-based Multi-\underline{v}iew Distributional R\underline{e}presentation \underline{L}earning (TraveL) framework, to capture both varied traveler behaviors and the regional correlations to learn path distributional representations. 
As shown in Figure~\ref{fig:TRAVEL-Framework}, the TraveL framework consists of a \textit{Path Encoder} to encode an input path, i.e., $p$, and a departure time (we consider \textit{rush hour} and \textit{normal hour} for simplicity) into a distributional representation, and an \textit{On-path Sequence Generator} which decodes a set of OP-Seqs from the path distributional representation to approach the ground truth, i.e., the historical OP-Seqs on $p$.
More specifically, Path Encoder first exploits the Road2Vec framework~\cite{fu2020trembr} to learn a latent vector as the initial embedding for each road segment.
The sequence of road segment embeddings in $p$, is then fed to a Multi-view Path Transformer, which applies regional attention to capture the regional correlation to output
a distribution representation for $p$, denoted as $N_p(\mu_p,\Sigma_p)$ (here we learn a Gaussian Distribution due to its generality).
From $N_p(\mu_p,\Sigma_p)$, we sample $n$ points, 
each of which is regarded as a representation of a possible OP-Seq. Each point is then fed to a Long-short Term Memory (LSTM) model to both generate the OP-Seq and recover $p$ with a series of predictions of the next road segments. 
To facilitate training, we measure the error for both OP-Seqs generation (i.e., travel time recovery) and the next road segment prediction (i.e., path recovery).
It is worth noting that we explore a novel idea of using K-S test~\cite{massey1951kolmogorov} to evaluate the error of OP-Seqs generation. 
Also note that, while the traditional variational autoencoder (VAE) also follows an encoder-decoder paradigm with a prior distribution included, TraveL and VAE are totally different. VAE learns to generate all the input paths from a prior distribution, where each path is generated from a latent vector sampled from the same prior. On the other hand, TraveL encodes each path to its own distributional representation in order to capture all possible traveler behaviors on the path.
Finally, to deal with the data sparsity issue of real-world datasets, we generate a synthetic dataset, Syn-Porto, for a complete evaluation of TraveL against several state-of-the-art models. 
We also conduct extensive experiments on two real-world datasets, i.e., Porto and Tokyo. The results show that TraveL outperforms all baseline models for three applications on all the datasets.


Major contributions made in this work are as follows.

\begin{itemize}
    \item We propose a novel idea of {\em distributional representation} to capture the varied traveler behaviors for PRL. 
    \item We analyze the \textit{regional correlation} in a path, and propose a \textit{regional attention} mechanism to capture it for PRL.
    \item We propose \textit{TraveL}, a new end-to-end framework, to capture varied traveler behaviors and regional correlation for PRL. We propose a novel sampling-based approach for varied traveler behavior generation, and exploit the idea of K-S test for the evaluation.
    
    
    
    

    \item We conduct extensive experiments on both synthetic and real-world datasets to evaluate TraveL against the state-of-the-arts PRL models.
    Experimental results show that \textit{TraveL} outperforms the state of the arts under various metrics for all three downstream applications, by reducing the mean K-S distance by 14.7\% for travel time distribution estimation.
\end{itemize}

\section{Related Work}\label{Related works}

We briefly review the related work on path representation learning and distributional representation learning.


\vspace{-0.1in}
\subsection{Path Representation Learning}

The goal of representation learning is to encode raw data into general-purpose low-dimensional latent vectors, i.e., embeddings, that are effectively fed as inputs to downstream machine learning and data mining methods for various applications. In recent years, neural network based representation learning methods have attracted a lot of research interests in various domains, e.g., text processing~\cite{mikolov2013efficient,le2014distributed,devlin2018bert,zhang2017active}, graph analytics~\cite{grover2016node2vec, fu2017hin2vec,sun2019infograph,goyal2020dyngraph2vec,xie2016representation}, computer vision~\cite{radford2015unsupervised,xia2014supervised,dosovitskiy2020image,chen2020uniter,lee2017unsupervised}, etc. 
Only recently, research on representation learning has been extended to path data, which are categorized into two folds. First, some existing works learn path representations for a specific targeted task, e.g., proximity search~\cite{liu2017semantic}, cascade prediction in a social network~\cite{li2017deepcas} and path ranking~\cite{yang2019pathrank}. The goal of these methods is not to learn general-purpose embeddings of paths, but is to exploit various recurrent neural network based models to aggregate the embeddings of road segments in a given path to generate a path representation for their specific. Second, aligned with our goal, some existing works learn general-purpose path embeddings. 
Among them, Yang et al. propose PathInforMax (PIM), with a curriculum negative sampling strategy to generate negative paths and two discriminators to distinguish the difference between the representation of the input path and those of its negative paths to learn path representations~\cite{yang2021unsupervised}.
Besides, BERT, a famous language model proposed by Delvin et al.~\cite{devlin2018bert} to capture long-term dependencies among nodes, is also examined by Yang et al. for PRL in their study.
However, BERT is not designed for road networks, failing to capture the regional correlation brought by road types. 
Fu et al. propose Trembr, by exploiting pre-trained road segment embeddings to capture static information in the road network. Trembr encodes an input path (i.e., a sequence of road segments) into its representation by using an LSTM-based encoder-decoder neural network structure~\cite{fu2020trembr}.
However, Trembr does not capture the varied traveler behaviors in the road network, and thus may not learn a proper path representations.


\subsection{Distributional Representation Learning}

Variational Autoencoder (VAE) is a famous framework that learns the mapping from a prior distribution to the distribution of the inputs~\cite{kingma2013auto,rezende2014stochastic}.
While VAE maps the input to a latent distribution, VAE is to learn a generator to generate outputs following the same distribution as the inputs from the prior, instead of learning a distributional representation for each input.
Recently, Ren et al. propose BETAE, which embeds entities and queries as Beta distributions, to capture the uncertainty of the queries, for multi-hop reasoning over knowledge graphs~\cite{ren2020beta}.
However, to the best of our knowledge, our work is the first attempt to learn path distributional representations in the road network.

\section{RESEARCH Problem and Challenges}\label{Preliminaries}
In this section, we introduce some key terms, define the targeted research problem and discuss the challenges.

\begin{definition}{\textbf{Road Network}}. 
A road network can be represented as a directed graph $G = (V,E,\Psi)$, where $V$ is a set of nodes representing intersections, associated with its coordinates $(v.lon, v.lat)$ (i.e., longitude and latitude); $E \subseteq V \times V$ is a set of directed edges representing road segments; and $\Psi : E \to \mathcal{F}$ is a function mapping an edge to its features, e.g., a road segment is a \textit{three-lane} \textit{highway}.
\end{definition}

As the traveler behaviors may be highly dependent on the departure time $d$ (in terms of rush hours or normal hours), we distinguish the two paths with the same road segments but different departure times. In other words, we define the term \textit{path} with $d$.

\begin{definition}{\textbf{Path}}. A path $p = \{r_1, r_2, r_3, ..., r_{|p|}; d\}$ is a sequence of connected road segments with a departure time $d$, where $r_i \in E$ is the $i$-th road segment in $p$, $|p|$ is the number of road segments in $p$, and $d$ is either \textit{rush hour} or \textit{normal hour}.
\end{definition}
\begin{definition}{\textbf{On-path Sequence}}. An on-path sequence $T = \{(r_1, t_{r_1}), (r_2, t_{r_2}), ..., (r_{|p|}, t_{r_{|p|}})\}$ consists of a path $p$ and the travel time spent on each road segment, where $r_i$ is the $i$-th road segment in $p$ and $t_{r_i}$ is the travel time on $r_i$.
\end{definition}

As the traveler behaviors are well captured in trajectory data, we formally define a trajectory as follows.  

\begin{definition}{\textbf{Trajectory}}. A trajectory is a sequence of spatio-temporal sample points, each of which contains a location (i.e., longitude and latitude) and a timestamp, generated from the movement of a traveler on a path.
\end{definition}

\noindent While trajectory (which is widely studied in many existing works) and path are related, please note the following difference to avoid confusion -- a path consists of a sequence of road segments, while a trajectory consists of a sequence of spatio-temporal sample points.

To observe and analyze the real-world traveler behavior on a path, i.e., travel time spent on a path, we project real-world trajectory data onto a road network using existing map mapping methods to obtain on-path sequences.
To capture the varied traveler behaviors and to increase the capacity of path representation, we explore to learn a \textit{distributional representation} for a path. We choose Gaussian
distribution as the form of path representations due to its generality. 
We define the research problem as below.

\begin{definition}{\textbf{Path Representation Learning in Road Networks}}. Given a dataset of selective paths $D=\big\{p_i\big\}_{i=1}^{|D|}$ in a road network $G$, where $p_i$ is the $i$-th path in $D$. The task of Path Representation Learning is to learn to map a path $p$ to a Gaussian distribution $N_p(\mu_p, \Sigma_p)$, where $\mu_p \in \mathbb{R}^d$ and $\Sigma_p \in\mathbb{R}^{d\times d}$, in a $d$-dimension space in support of
a variety of path mining tasks. 
\end{definition}



In our work, we explore encoder-decoder paradigm, which first encodes various signals on the targeted path, with the regional correlation captured, into the distributional representation, and then decodes the representation to generate varied traveler behaviors. 
To realize the idea, we face the following challenges: (1) \textit{Recovery of the varied traveler behaviors}. The travel time spent on the road segments by travelers along a path may not follow a known distribution (e.g., Gamma or Chi-squared distribution).
Thus we are not able to statistically fit the historical travel times to a known parameterized distribution as the path representation.
How to reasonably generate varied traveler behaviors, i.e., varied travel times, from the distributional representation?
How to evaluate the generated travel times with the ground truth?
How to deal with the data sparsity issue in real-world datasets, i.e., the lack of ground truth travel times on each path?
(2) \textit{Capturing the regional correlation in the path.} Existing models fail to capture regional correlation brought by various road segment relationships.
What road segment relationships bring the regional correlation, that should be captured? How to capture regional correlation with Path Encoder? In the following, we introduce our design of TraveL to address these questions.

\section{Design of TraveL}\label{Methods}
In this section, we first introduce the proposed TraveL framework and then detail our design of its components.

\vspace{-0.05in}
\subsection{The TraveL Framework}
To capture both the varied traveler behaviors and the regional correlation on the path for PRL, we follow the encoder-decoder paradigm in the design of TraveL. As shown in Figure~\ref{fig:TRAVEL-Framework}, TraveL includes three components: 1) a \textit{Path Encoder} to encode the sequence of road segments in the target path $p$, together with the departure time, to its distributional representation in the latent space; 2) an \textit{On-path Sequence (OP-Seq) Generator} which generates a set of possible OP-Seqs on $p$, which are expected to follow the same distribution in travel behaviors (i.e., travel times) as the ground-truth, i.e., the historical OP-Seqs on $p$, and recovers $p$ by sequentially predicting the next road segment; and 3) an \textit{On-path Sequence Evaluator} to evaluate both the recovery of the path and the generation of OP-Seqs to facilitate training. 
Here, we propose a sampling-based approach to generate the OP-Seqs: we regard a sample point from the path distributional representation as the representation of a possible OP-Seq.
As such, the distributional representation covers all possible OP-Seqs on the path and thus acts as a proper path representation with the varied traveler behaviors well captured.


Specifically, given a path $p$ =  $\{r_1, r_2, ..., r_{|p|}\}$, $p$ is fed to the Path Encoder to generate its {\em distributional representation} followed Gaussian distribution, denoted as $N(\mu_p, \Sigma_p)$. 
A sampling process is then applied to sample $n$ points, i.e., $s_1, s_2, ..., s_n$, from $N_p(\mu_p, \Sigma_p)$. 
After that, the sample point $s_j$, is fed to the OP-Seq Generator to generate an OP-Seq $T^{\prime}_j=\{(r_1, {t^{\prime}}^j_{r_1}), ..., (r_{|p|}, {t^{\prime}}^j_{r_{|p|}})\}$, where  ${t^{\prime}}^j_{r_k}$ is the generated travel time on $r_{k}$. Meanwhile, the OP-Seq Generator predicts the next road segment, i.e., $r^\prime_{k+1}$, given the partial path till $r_{k}$.
With these outputs fed,
to evaluate the generated travel times, OP-Seq Evaluator collects historical OP-Seqs on $p$, and measure the dissimilarity between the set of generated travel times and the set of historical travel times with K-S distance. Besides, OP-Seq Evaluator treats the next road segment prediction as a classification task, with a maximum likelihood based loss function applied. With both the paths and historical OP-Seqs fed, we train TraveL end to end.
Owing to the proven power of these loss functions, we leave the idea of learning a neural net based OP-Seq Evaluator with adversarial learning in future.
Note that, to simplify the sampling process, we assume that the learned Gaussian distribution has $\Sigma_p = {\sigma_p}^2 I$. Owing to the power of the OP-Seq Generator to learn complicated transformation from $N_p(\mu_p, {\sigma_p}^2I)$ to the historical OP-Seqs, this assumption (and the adoption of Gaussian distribution as the path distributional representation) may not compromise the model's ability to recover Op-Seqs.


\vspace{-0.05in}
\subsection{Path Encoder}
\begin{figure}[t]
\centering
\includegraphics[width=2.8in]{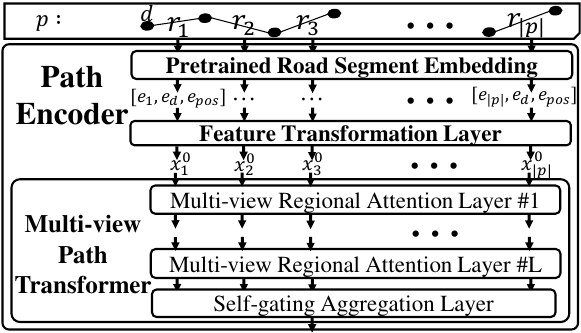}
\caption{The Structure of Path Encoder}
\centering
\label{fig:path-encoder}
\end{figure}

In this section, we detail our design of Path Encoder, which encodes a path and its departure time to a distributional representation. 
As shown in Figure~\ref{fig:path-encoder}, Path Encoder first encodes each road segment to an embedding with a process of pre-training and feature transformation, and then feed the new road segment embeddings to a Multi-view Path transformer to capture the regional correlation to enhance road segment embeddings, and finally aggregate the road segment embeddings to generate the path distributional representation. Now, we detail all the steps.



\subsubsection{Road Segment Embedding Initialization.}
Owing to the proven power of Road2Vec for road segment embedding pre-training~\cite{fu2020trembr}, in this work, we first exploit Road2Vec to generate the initial road segment embeddings, denoted as $\{e_i, i=1,2,...,|p|\}$ for $p$. 
In addition, we exploit a two-dimension one-hot departure time embedding, $e_d$, which is $(0,1)^T$ when $d$ is in rush hour (8-10AM and 5-7PM), and $(1,0)^T$ when $d$ is in normal hours or $(0,0)^T$ when $d$ is not known.
Then we exploit a one-layer feed-forward net (FFN) to transform the concatenation of $e_i$, $e_d$ and a positional embedding $e_{pos}$ (i.e., a sinusoidal function used in Transformer~\cite{vaswani2017attention}) to a same latent space,  i.e., $
 x^0_i = FFN([e_i, e_t, e_{pos}]) = W_2 ReLU(W_1 [e_i, e_t, e_{pos}]+b_1) + b_2$, 
where $W_1$, $W_2$ are two feature transformation matrices; $b_1$, $b_2$ are two bias vectors; and ReLU is the non-linear activation function. 

\begin{figure}[!htbp]
\vspace{-0.07in}
\centering
\includegraphics[width=2.2in]{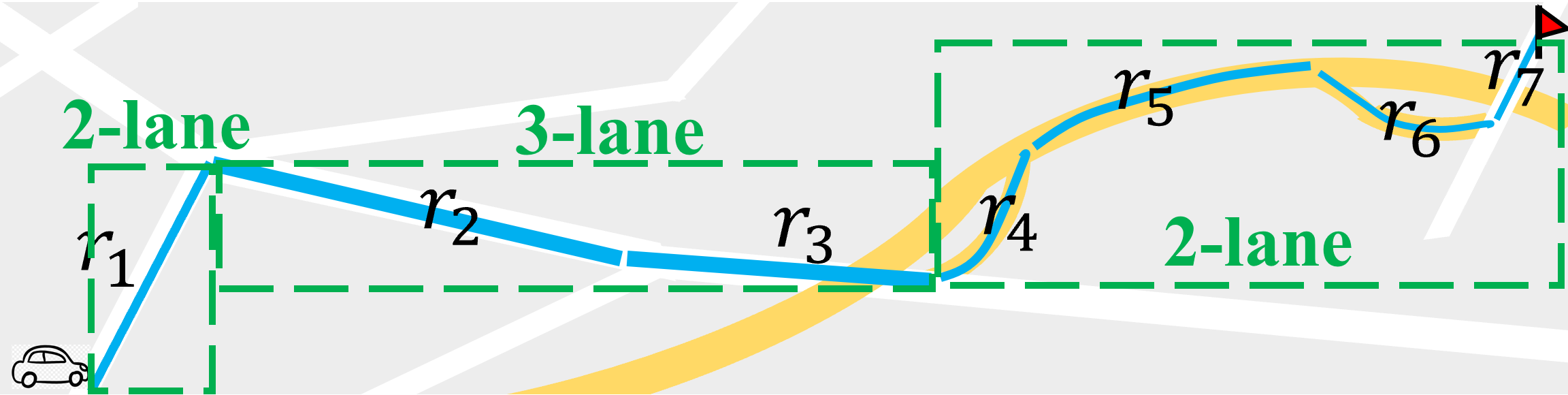}
\vspace{-0.1in}
\caption{Regional Correlation of a Path in Lane View}\centering
\label{fig:path-lane}
\vspace{-0.2in}
\end{figure}

\subsubsection{Multi-view Path Transformer}
Here we first analyze the regional correlation in the path.
We use the term \textit{path region} to denote some sequence of consecutive road segments in the path, where the traveler behaviors within a path region are correlated.
We observe two types of road segment relationship bringing the regional correlation: i) same road-type relationship. The traveler behaviors in a region of the same road type, may be correlated. Figure~\ref{fig:path-lane} shows an example where travelers start the trip from the left-bottom corner and the destination is at the right-upper corner. We observe that $r_2$ and $r_3$ are three-lane roads, while the rest are two-lane roads. Intuitively, the traveler behavior may change significantly when the traveler drives from a two-lane road to a three-lane road, while keep stable on a consecutive sequence of two-lane road segments; and ii) co-occurrence relationship. Traveler behaviors within a number of hops along the path, may be correlated.

Thus, we propose to define different \textit{views} of the paths, where each view denotes a path structure exhibiting some specific regional correlation, e.g., the lane view which has three path regions (as shown in Figure~\ref{fig:path-lane} with green bounding boxes). More generally, Algorithm \ref{alg:type-view} shows the process to generate path regions under a type-based view for the type $\mathcal{T}$. 
In this work, we explore the \textit{highway view} and the \textit{lane view} as the representative type-based views.
In addition, we propose a \textit{hop view}, where every $H$ (a hyper-parameter to decide) continuous road segments in the path form a region, i.e., $\{r_1,...,r_H\}$ as the first region and $\{r_{H+1},...,r_{2H}\}$ as the second region, etc.
However, co-occurrence relationships among road segments around the region borders, e.g., $r_H$ and $r_{H+1}$, are not shown as they are in different regions.
To resolve the issue, we explore the idea of \textit{region shifting} to shift all regions right together, by one road segment, to form a new view. As such, with at most $(H-1)$ times shift, each pair of two correlated road segments are guaranteed to be in the same region at least in one view. 
All the $(H-1)$ views work together to exhibit the co-occurrence relationships.


\setlength{\textfloatsep}{0.02cm}
\begin{algorithm}[t]
 \caption{Region generation under a type-based view}
 \label{alg:type-view}
 \begin{algorithmic}[1]
\REQUIRE input path $p={r_1, r_2, ..., r_{|p|}}$; road segment type values $\mathcal{T}_1, \mathcal{T}_2, ..., \mathcal{T}_{|p|}$ of a given type $\mathcal{T}$; 
\STATE Initialize the set of regions $\mathcal{R}=\varnothing$, road segment index (to assign) $i=1$, region (to generate) index $k=1$;
\REPEAT
\STATE Initialize $\mathcal{R}_k=\varnothing$
\WHILE{$i\leq |p|$ and ($\mathcal{R}_k$ is $\varnothing$ or ($r_{i-1} \in \mathcal{R}_k$ and $\mathcal{T}_{i-1} = \mathcal{T}_{i}$))}
\STATE $\mathcal{R}_k$ = $\mathcal{R}_k \cup \{r_i\}$, $i=i+1$
\ENDWHILE
\STATE $\mathcal{R}=\mathcal{R}\cup \{\mathcal{R}_k\}$, $k=k+1$
\UNTIL $i>|p|$
\end{algorithmic}
\end{algorithm}
\setlength{\floatsep}{0.02cm}

\begin{figure}[b]
\centering
\includegraphics[width=2.6in]{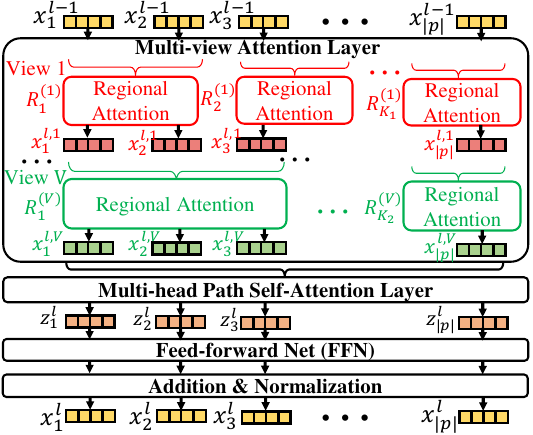}
\vspace{-0.1in}
\caption{The $l$-th Multi-view Regional Attention Layer}
\vspace{-0.3in}
\label{fig:mvp-encoding-layer}
\end{figure}

To capture the regional correlation under each view, we propose the idea of \textit{regional attention}, which correlates each road segment to other road segments in the same region. 
We implement regional attention in each of the stacked Multi-view Regional Attention (MVRA) Layers in the Multi-view Path Transformer, and Figure~\ref{fig:mvp-encoding-layer} shows the structure of the $l$-th MVRA layer.
Generally, road segment embeddings are first fed to a Multi-view Attention Layer to capture the regional correlation to generate a new embedding for each road segment.
Then the new embeddings generated from different views are averaged and fed to a Multi-head Path Self-attention Layer to capture the long-term dependency among the road segments. Finally, the road segment embeddings are fed to an FFN for a non-linear transformation, and then fed with a residual connection and a layer normalization to be the road segment representations input to the $(l+1)$ MVRA layer. 

More specifically, we denote the embedding of $r_i$ input to the $l$-th MVGA layer as $x^{l-1}_i$. Under a view $v$, the path is split to $K_v$ regions, which are denoted as $R^{(v)}_1$, ..., $R^{(v)}_{K_v}$ respectively.
Then, for the $k$-th region, i.e., $R^{(v)}_{k}$, and each road segment $r_i \in R^{(v)}_{k}$, the regional attention is to generate a new embedding, denoted as $x^{l,v}_{i}$, for $r_i$ by capturing its correlation with other road segments in the region with the following steps: (1) we apply feature transformation on each $x^{l-1}_i$ $\in$ $R^{(v)}_k $ with a query matrix $W^{l,v}_Q$, an answer matrix $W^{l,v}_A$ and a value matrix $W^{l,v}_H$ to transform $x^{l-1}_i$ to its query vector $q^{l,v}_i$, answer vector $a^{l,v}_i$ and value vector $h^{l,v}_i$ by $    q^{l,v}_i=W^{l,v}_Q\cdot x^{l-1}_i$, $a^{l,v}_i=W^{l,v}_A\cdot x^{l-1}_i$ and $h^{l,v}_i=W^{l,v}_H\cdot x^{l-1}_i$, respectively. 
(2) for each $r_j \in R^{(v)}_k$, we propose to derive the impact of $r_j$ on $r_i$ by calculating an attention score, $\alpha^{l,v}_{ij} = f^{l,v}_\alpha(q^{l,v}_i, a^{l,v}_j) = ({q^{l,v}_i})^T W^{l,v}_{\alpha} a^{l,v}_j$, 
where $f^{l,v}_\alpha(\cdot,\cdot)$ is the attention function with $W^{l,v}_{\alpha}$ as a feature transformation matrix to learn. (3) we normalize the attention scores with a softmax function, and exploit the normalized attention scores to aggregate the value vectors to generate $x^{l,v}_{i}$ as follows,
\setlength\abovedisplayshortskip{-6pt}
\begin{equation}\small
    \beta^{l,v}_{ij} = \frac{exp(\alpha^{l,v}_{ij})}{\sum_{r_j\in R^{(v)}_k} exp(\alpha^{l,v}_{ij})},\quad
x^{l,v}_i = \sum_{r_j\in R^{(v)}_k} \beta^{l,v}_{ij} h^{l,v}_j
\end{equation}
\setlength\belowdisplayshortskip{0pt}
The generated embeddings of $r_i$ under different views, i.e., $x^{l,1}_i$, ..., $x^{l,V}_i$, are then averaged to be ${x'}^{l}_i$ to feed to the following Multi-head Path Self-Attention Layer, which follows the attention design in Transformer~\cite{vaswani2017attention}, to capture the long-term dependency among road segments to output $z^{l}_i$ as the embedding of $r_i$.
Finally, $z^{l}_i$ is fed to a two-layer FFN for further feature transformation, 
and then fed to a layer-normalization layer with a residual link built to generate $x^l_i$, i.e., $x^l_i = LayerNorm({FFN(z^l_i)}+x^{l-1}_i)$.
By stacking $L$ MVRA layers, road segment representations have more opportunity to correlate with other road segments to capture more complicated relationships. At the end, $\{x^{L}_i, i=1,2,...,|p|\}$ is fed to a Self-gating Aggregation Layer to generate the path distributional representation.

For aggregation, the idea is to first transform the embedding of $r_i$ into two embeddings in different latent spaces, i.e., $x^\mu_i$ and $x^\sigma_i$, with the former for prediction $\mu_p$ and the latter for prediction $\sigma_p$; and then exploit a weighted sum of $\{x^\mu_i\}$ as $\mu_p$ and a weighted sum of $\{x^\sigma_i\}$ as $\sigma_p$.
Formally, we generate $x^\mu_i$ and $x^\sigma_i$ by, $x^\mu_i=tanh(W_{\mu}x^L_i+b_\mu)$ and $x^\sigma_i=tanh(W_{\sigma}x^L_i+b_\sigma)$.
Then we adopt a self-gating mechanism to derive the weight of each road segment to the path, and normalize the weights as follows,
\begin{equation}
\gamma^{\mu}_i = \frac{exp(f_\mu(x^\mu_i))}{\sum_i exp(f_\mu(x^\mu_i))}, 
\gamma^{\sigma}_i = \frac{exp(f_\sigma(x^\sigma_i))}{\sum_i exp(f_\sigma(x^\sigma_i))} 
\end{equation}
where $f_\mu(\cdot,\cdot)$ and $f_\sigma(\cdot,\cdot)$ are two two-layer FFNs with output dimension as 1 to derive the weights and a Softmax function is used for normalization.
At the end, we aggregate the road segment embeddings with the normalized weights to be the path distributional representation by
$\mu_p=\sum^{|p|}_{i=1}\gamma^\mu_ix^\mu_i$ and  $\sigma_p=\sum^{|p|}_{i=1}\gamma^\sigma_ix^\sigma_i
$.
Next, the path distributional representation, $N(\mu_p, {\sigma_p}^2I)$, is fed to an On-Path Sequence Generator to generate possible OP-Seqs on $p$.

\subsection{On-path Sequence Generator}
\begin{figure}[t]
\vspace{-0.1in}
\centering
\includegraphics[width=2.8in]{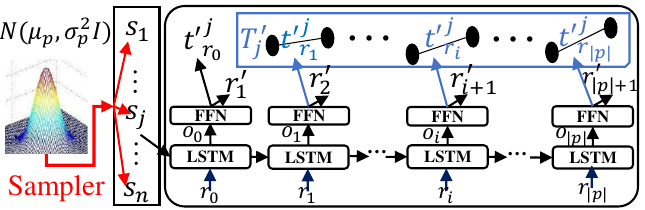}
\vspace{-0.1in}
\caption{Structure of On-path Sequence Generator}\centering
\label{fig:on-path-sequence-generator}
\end{figure}

As shown in Figure~\ref{fig:on-path-sequence-generator}, the OP-Seq Generator generates a set of possible OP-Seqs from $N(\mu_p, {\sigma_p}^2I)$ while recovering $p$. 
To generate OP-Seqs, $n$ points, i.e., $\{s_j, j=1,2,...,n\}$, are sampled~\footnote{To facilitate training, we follow VAE to exploit a reparameterization trick.} from $N(\mu_p, {\sigma_p}^2I)$. 
Then $s_j$ is fed to a Long Short Term Memory (LSTM), a commonly used sequential data generator, to jointly generate the travel time spent on each road segment (marked as blue in Figure~\ref{fig:on-path-sequence-generator}) and predict the next road segments one by one.
At time point 0, $r_0$, which is always a START road segment telling the LSTM to start the generation, is fed to the LSTM to predict the first road segment, i.e., $r^\prime_1$.
Then at time step $i$, LSTM takes $r_i$ as the input and output $o_i$ as a summarization of the partial path $\{r_1, r_2, ..., r_i\}$. $o_i$ is then fed to two FFNs, respectively, to generate: 1)  ${t^{\prime}}^j_{r_i}$, i.e., the travel time spent on $r_i$; and ii) the probability of each road segment to be the next road segment with a softmax function applied.
As a special case, we do not care about ${t^{\prime}}^j_{r_0}$ since it is not a meaningful travel time on any road segment. Meanwhile, we do maximize the predicted probability of the END road segment to be $r^\prime_{|p|+1}$.

\subsection{Loss Design in On-path Sequence Evaluator}
In this section, we detail the design of our loss function in the OP-Seq Evaluator.
Note that the output of OP-Seq Generator includes two parts: i) the generated OP-Seqs, i.e., $\{T^\prime_j|j=1,2,...,n\}$
where $n$ is the number of sample points; 
and ii) the predicted probability distribution of the next road segments. We design loss functions for them respectively.

\noindent\textbf{Loss for OP-Seqs Generation.}
Given an input path $p$, we let ${S}^\prime_p = \{{t^\prime}^1_p, ..., {t^\prime}^n_p\}$ denote the travel times spent on the $n$ generated OP-Seqs, i.e., ${t^\prime}^j_p=\sum^{|p|}_{i=1} {t^\prime}^j_{r_i}$. 
Suppose that $p$ has $m$ historical OP-Seqs. Let the set of travel times spent on these historical OP-Seqs be denoted as $S_p = \{t^{1}_p, t^{2}_p, ..., t^{m}_p\}$.
To measure the distance between $S^\prime_p$ and $S_p$, we exploit the two-sample Kolmogorov–Smirnov (K-S) test, which tests the equality of two continuous one-dimensional probability distributions by measuring the distance, called \textit{K-S distance}, between two sets of values sampled from the two distributions respectively. 
Thus in TraveL, we propose to exploit the K-S distance as the loss function for OP-Seq generation, i.e.,
\begin{equation}
\mathcal{L}_{PathTime}(\theta, p)=KS(S^\prime_p(\theta), S_p)
\end{equation}
where $\theta$ points to the TraveL model parameters and $S^\prime_p(\theta)$ is the $S^\prime_p$ generated with $\theta$, and KS($\cdot,\cdot$) is the K-S distance function (see Section 5.8).\\
\noindent \textbf{Data Sparsity Issue.} We observe the data sparsity issue in collected real-world data: the number of historical OP-Seqs on a path $p$, i.e., $|S_p|$, may be so small, that there may be a large bias between the observed $S_p$ and the real travel time distribution on $p$, rendering the K-S distance defined above not working.
To address the issue, we propose to apply a regularization on the travel time generation, by asking the generated travel time distribution on each road segment $r_i \in p$ to also approach the historical travel time distribution on $r_i$.
Denote the set of generated travel time on $r_i$ as $S^\prime_{r_i}(\theta) = \{{t^\prime}^j_{r_i}| j=1,2,...,n\}$ with TraveL model parameters $\theta$, and the set of historical travel time (not necessary to be on $p$) on $r_i$ as $S_{r_i}$. 
We define the loss for road segment travel time generation as the K-S distance between $S^\prime_{r_i}$ and $S_{r_i}$, i.e., $\mathcal{L}(\theta, r_i)=KS(S^\prime_{r_i}(\theta), S_{r_i})$.
We sum the loss for all road segments in path $p$ as follows.
\begin{equation}\small
    \mathcal{L}_{RsTime}(\theta, p) =\sum_{r_i\in p} \mathcal{L}(\theta, r_i) = \sum_{r_i\in p} KS(S^\prime_{r_i}(\theta), S_{r_i})
\end{equation}
which acts as a regularization term to supplement $\mathcal{L}_{PathTime}(\theta, p)$.


\noindent\textbf{Loss for the Next Road Segment Prediction.}
In addition to evaluation on the generated OP-Seqs, we also evaluate the prediction of the next road segment. 
We treat the next road segment prediction task as a classification problem, and train the model by maximizing the probability of the ground truth next road segment to be the next road segment at each time step of prediction.
Thus, we explore a maximum likelihood based loss function for the next road segment prediction as follows.
\begin{equation}
    \mathcal{L}_{RsPred}(\theta, p) =
    - \sum^n_{j=1} \sum^{|p|}_{i=0} log P(r_{i+1}|s_j, r_0, r_1, ..., r_i, \theta)
\end{equation}

Finally, we regularize the learned distributional representation $N$($\mu_p, \sigma^2_p I$) with $\mathcal{L}_{PathRep}(\theta, p) = KL(N(\mu_p, \sigma^2_p I), N(0,I)))$, and the model parameters $\theta$ with a two-norm regularization. 
We sum these losses for all paths in the dataset $D$ to get the final loss function as follows.
\begin{equation}\small
\begin{split}
\mathcal{L}(\theta) = \frac{1}{|D|}\sum_{p\in D}( \lambda_1 \mathcal{L}_{PathTime}(\theta, p) + \lambda_2 \mathcal{L}_{RsTime}(\theta, p)\\ + \lambda_3 \mathcal{L}_{RsPred}(\theta, p) + \lambda_4 \mathcal{L}_{PathRep}(\theta,p))
+ \lambda_5{\lvert\lvert \theta \rvert\rvert}^2_2
\end{split}
\end{equation}
where $|D|$ is the dataset size and $\sum^3_{i=1} \lambda_{i}=1$ to weight the losses.

\section{Performance Evaluation}\label{Experiments}
In the following, we introduce the datasets, baseline models for comparison, and a typical representation learning process.
Then we evaluate TraveL distributional representations against baselines on three tasks, i.e., travel time distribution estimation, path similarity prediction and destination prediction, for all the datasets.


\subsection{Datasets}
We collect two commonly used real-world trajectory datasets, Porto and Tokyo, and generate a synthetic dataset, Syn-Porto, on the map of Porto for the evaluation. 
We briefly introduce the datasets below.



\noindent\textbf{Porto} collects 1.7 minion taxi trajectories of 442 taxis in Porto, Portugal from January 2013 to June 2014\footnote{http://www.
geolink.pt/ecmlpkdd2015-challenge/.}. 

\noindent\textbf{Tokyo} collects 78 million GPS sample points from 617K users walking or taking vehicles in Tokyo~\cite{kashiyama2017open}. We extract the GPS point sequences for users taking bike or car, and then segment the sequences into trajectories with a 45-second gap.

\noindent \textbf{Syn-Porto} is the synthetic dataset including 200 million OP-Seqs of moving taxis in Porto, generated by a simulation system. 
Because that we have no information about the real travel time distribution for synthesis, we create three versions of Syn-Porto, denoted as $D_{nor}$, $D_{logn}$ and $D_{mix}$ respectively, with road segment travel times sampled from a Normal distribution, a Log-normal distribution, or a mixture of them.
The three versions are used to prove the consistent superiority of TraveL under various assumptions.

For Porto and Tokyo, we remove trajectories with less than 10 GPS sample points (too short) and finally obtain 1.2 million and 0.29 million trajectories, respectively. 
To obtain OP-Seqs, we adopt Barefoot\footnote{Barefoot can be found at https://github.com/bmwcarit/barefoot/.}, a Hidden
Markov Model based map matching tool, to project a trajectory onto a
road network to yield an OP-Seq.
We further exploit the three-sigma principle to remove OP-Seqs with outlier road segment travel time.

\subsection{Baseline Models}
In this section, we introduce the state-of-the-art path representation learning models 
compared in the evaluation.

\noindent \textbf{Node2Vec} learns node presentations in a graph~\cite{grover2016node2vec}. We average the representations of nodes along a path as the path representation.

\noindent \textbf{RoadSegment (RS)} feeds the sequence of road segments in the path to a seq2seq model, where the encoder
and decoder are both a single layer LSTM model, to generate the path representation.

\noindent \textbf{InfoGraph} is an unsupervised model to learn a representation for the whole graph~\cite{sun2019infograph}. In this work, we regard a
path as a special graph and learn the path representation with InfoGraph.

\noindent \textbf{BERT} is an unsupervised language representation learning model~\cite{devlin2018bert}. 
We treat a road segment as a word and a path as a sentence. To enable training, we split a path P into two sub-paths P1 and P2 (with equal number of road segments), and consider (P1, P2) as a valid pair of two continuous sentences and (P2, P1) as invalid because the latter does not form a real path in the road network.

\noindent \textbf{Path InfoMax (PIM)} is an unsupervised path representation learning framework which first generates negative samples with curriculum negative sampling, and then employs mutual information maximization to learn path representations~\cite{yang2021unsupervised}.

\noindent \textbf{Trembr} is a state-of-the-art trajectory representation learning model which first maps a trajectory to a path on road network to learn the trajectory representation~\cite{fu2020trembr}. To learn a trajectory representation, Trembr feeds the sequence of road segments with pre-trained embeddings into a seq2seq model with a special designed loss function to distinguish relationships among road segments. In this work, we adapt Trembr to learn the path representation.

\noindent \textbf{TraveL} feeds the sequence of road segments into the proposed TraveL framework to generate path distributional representations.

\vspace{-0.03in}
\subsection{Representation Learning Process}
Given a dataset $D$ consisting of a set of paths and their on-path sequences, we randomly split the set of all paths into 90\% and 10\% as training and validation set, respectively. We train the TraveL model with the paths in the training set and choose the optimal model parameters which achieve the lowest loss on the validation set. The initial learning rate is set to 0.0001 and we use Adam optimizer for training. We further tune the model hyper-parameters to minimize the loss.
The optimal hyper-parameter setting is detailed in Section
5.9. 
We also tune the best parameter settings for all baseline models. The dimensionality of path representations learned by baseline models is finally set as 256. 

\subsection{Travel Time Distribution Estimation (TTDE)}
To evaluate all the PRL models for the TTDE task, we first learn the path representations for each model following the training process in Section 5.3. For each dataset, once we learn the path representations, we randomly split all the paths, associated with their learned representations, into 80\%, 10\% and 10\%, as the training, validation and testing set, respectively, for the TTDE task.
For each path in the training set, we randomly sample a set of points from its distributional representation, and feed these points to a two-layer FFN to generate a set of travel times, respectively. 
We train the FFN by minimizing the K-S distance between the generated travel times and the ground truth, with the validation sets to select the optimal FFN.
Finally, we evaluate the TraveL representations on the testing set with K-S distance between the generated (with the trained FFN) and ground-truth travel time set as the metric. However, one issue is that, for baseline models, each path representation is a vector, which cannot be used for sampling.
To fill the gap, we train another two FFNs to first map the vector to a pair of 128-dim mean vector and standard deviation vector, and then follow the sampling process introduced above to train all FFNs with the K-S loss end to end.


For Porto and Tokyo, due to the data sparsity issue, we evaluate the representations of sub-paths (of a fixed length $L$), which have sufficient OP-Seqs in the real-world datasets.
We collect the sub-paths and their OP-Seqs, and train all models to obtain sub-path representations following the representation learning process introduced in Section 5.3.
Then we evaluate sub-path representations, in a same way as full path representations. We vary $L$ as 5, 10, 15 and `Full' (where we learn representations for full paths) to evaluate the impact of $L$ to the model performance.

\begin{table}[t]
\caption{Travel Time Distribution Estimation on Syn-Porto}
\vspace{-0.15in}
\begin{center}
\resizebox{\linewidth}{!}{ 
\begin{tabular}{|c|c|c|c|c|c|c|c|c|c|}
\hline
\multicolumn{1}{|c|}{ }&\multicolumn{3}{c|}{$D_{mix}$}&\multicolumn{3}{c|}{$D_{nor}$}&\multicolumn{3}{c|}{$D_{logn}$}\\
\hline
\multicolumn{1}{|c|}{Model }&\multicolumn{1}{c|}{Short}&\multicolumn{1}{c|}{Mid}&\multicolumn{1}{c|}{Long}&\multicolumn{1}{c|}{Short}&\multicolumn{1}{c|}{Mid}&\multicolumn{1}{c|}{Long}&\multicolumn{1}{c|}{Short}&\multicolumn{1}{c|}{Mid}&\multicolumn{1}{c|}{Long}\\
\cline{1-10}
Node2Vec&
0.31 & 0.34 & 0.38& 
0.28 & 0.33 & 0.37& 
0.27 & 0.30 & 0.36\\
\cline{1-10}
RS&
0.32 & 0.36 & 0.39 & 
0.30 & 0.34 & 0.38& 
0.29 & 0.32 & 0.38\\
\cline{1-10}
BERT &
0.38 & 0.40 &0.42& 
0.36& 0.39 & 0.41& 
0.36& 0.38 & 0.40\\
\cline{1-10}
InfoGraph &
0.35 & 0.37 & 0.40& 
0.33 & 0.36 & 0.42& 
0.32 & 0.34 & 0.39\\
\cline{1-10}
PIM &
0.29 & 0.32 & 0.35$^*$& 
0.26$^*$ & 0.30 & 0.35& 
0.26 & 0.29 & 0.35\\
\cline{1-10}
Trembr &
0.28$^*$ & 0.31$^*$ & 0.35$^*$& 
0.26$^*$ & 0.28$^*$ & 0.33$^*$& 
0.25$^*$ & \textbf{0.28}$^*$ & 0.34$^*$\\
\cline{1-10}
TraveL &
\textbf{0.24} & \textbf{0.29} & \textbf{0.33}& 
\textbf{0.23} & \textbf{0.27} & \textbf{0.31}& 
\textbf{0.23}& \textbf{0.28} & \textbf{0.29}\\
\hline
\end{tabular}
}
\label{syn-travel-time-distribution-estimation}
\end{center}
\end{table}

\noindent \textbf{Evaluation of Models} 
Table~\ref{syn-travel-time-distribution-estimation} shows the Mean K-S distance (MKS) achieved by different models on the three synthetic datasets.
To analyze the impact of the path length to the model performance, we categorize all paths to Short (<3km), Mid (3km-6km), or Long (>6km), and show the MKS achieved for paths of each category.
As shown, TraveL outperforms all the baselines consistently, for various lengths on all three synthetic datasets, by reducing MKS from 3.57\% (0.27 v.s. 0.28 by Trembr on $D_{nor}$) to 36.1\% (0.23 v.s. 0.36 by BERT on $D_{logn}$). We observe that model performance on $D_{mix}$ is worse than $D_{nor}$ and $D_{logn}$, which fits our intuition that a more complicated travel time distribution is harder to predict. Among the baselines, we observe that PIM and Trembr outperform others because PIM distinguishes the co-occurrence relationships among nodes and paths, and Trembr captures same-type relationship among the road segments as well into the pre-trained road segment embeddings.

In addition, we observe all models perform worse (i.e., achieve higher MKS) for longer paths.
An interesting observation is that, for most models, the increase of MKS from Mid to Long is larger than that from Short to Mid (e.g., MKS by Trembr increases 0.05 from Mid to Long v.s. 0.02 from Short to Mid), which indicates that these models get worse more rapidly for longer paths.
The challenge for long path travel time distribution prediction comes from both the dependency capturing for road segments in a longer distance and the more complicated travel time distribution (since it is a sum of more road segments' travel time distribution). Instead, MKS by TraveL increases 0.04 on $D_{mix}$ from Mid to Long and 0.05 from Short to Mid, which indicates that TraveL works well for learning representations for long paths owing to its ability to capture long-term dependencies and the varied traveler behaviors. 
\begin{figure}[t]
\centering
\includegraphics[width=3in]{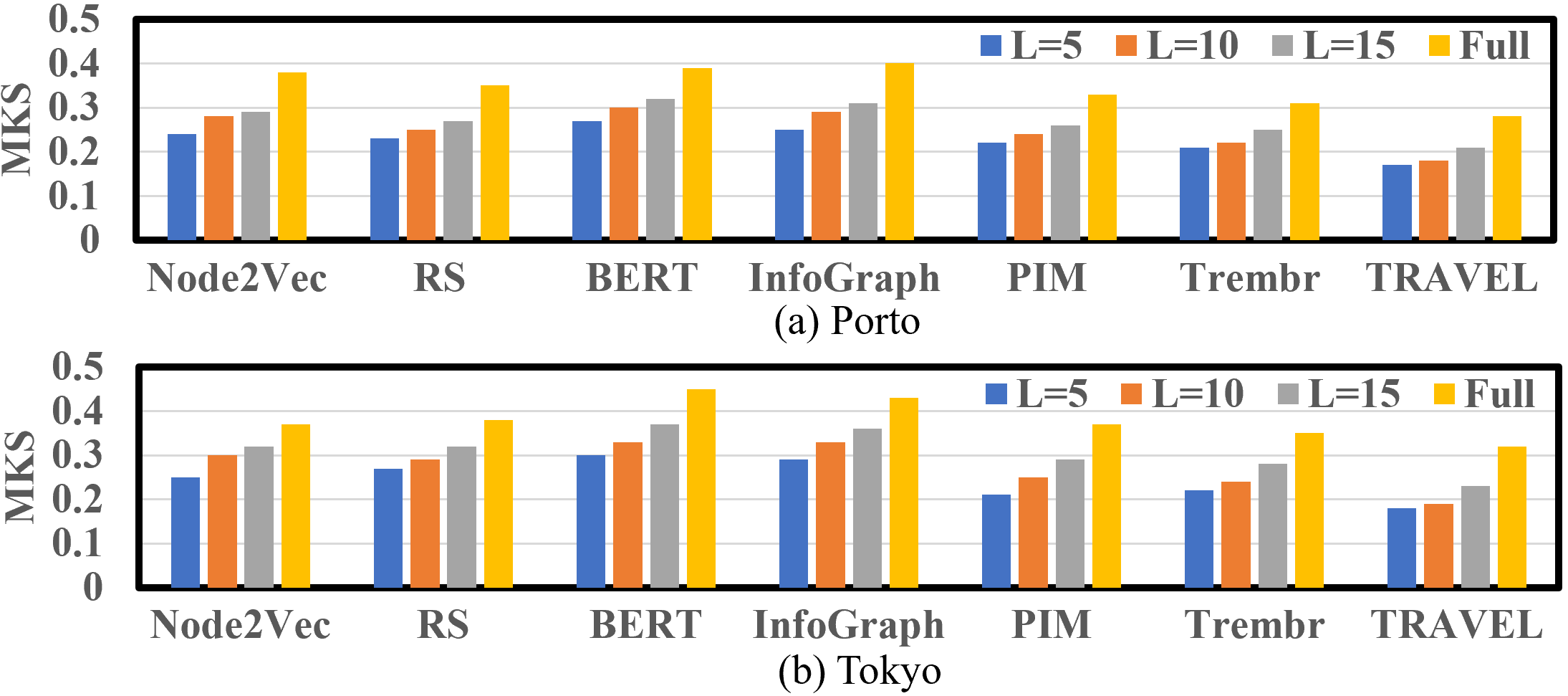}
\vspace{-0.15in}
\caption{Travel Time Estimation on Porto and Tokyo}
\label{fig:real-travel-time-estimation}
\end{figure}
On Porto and Tokyo, we observe that TraveL also outperforms the baseline models for sub-paths of all lengths (shown in Figure ~\ref{fig:real-travel-time-estimation}).
As shown, all models perform worse with a larger path length ($L$), which suggests the challenge brought by long paths. 
We observe that, BERT, an baseline that also exploits the Transformer structure to capture the long-term dependencies, does not work as well as TraveL. The reason may be that TraveL also captures the regional correlation with the regional attention mechanism and the varied traveler behaviors, which are both essential for the TTDE task.


\vspace{-0.03in}
\subsection{Path Similarity Prediction (PSP)}
Here we learn path representations with the representation learning process introduced in Section 5.3 for the PSP task. 
The learned path representations are believed to capture rich information if they can be used to predict various kinds of path similarity.
In this experiment, we target on the following two similarities: i) the weighted Jaccard similarity (WJ), a commonly used path similarity function, which is the ratio of the shared road segments' length to the total length of the two paths; and ii) the speed-relevant weighted Jaccard similarity (SWJ) we propose, 
which is the ratio of the expected travel time spent on the shared road segments to the expected total travel time spent on the two paths. We define SWJ between two paths, $p^{(1)}$ and $p^{(2)}$, as follows.
\begin{equation}
    SWJ(p^{(1)}, p^{(2)}) = \frac{ \sum_{r\in p^{(1)}}{\textbf{1}_{[r\in p^{(2)}]} ET(r)} }{\sum_{r\in p^{(1)}}ET(r) +\sum_{r\in p^{(2)}}ET(r)  }
\end{equation}
where $\textbf{1}_{[r\in p^{(2)}]}$ is an indicator function which equals to 1 when $r\in p^{(2)}$ and equals to 0 otherwise, $ET(r)$ is the expected travel time on $r$, i.e., the length of $r$ divided by the speed limit on $r$.
Intuitively, predicting SWJ is more challenging (than WJ), since it requires to capture speed-relevant information in addition to lengths.

Next, we generate a dataset $D_{pp}$ consisting of path pairs, from each path dataset. 
For Syn-Porto, given a path $p$, we follow an existing literature, PathRank, to collect candidate paths with the same origin and destination as $p$, which are then paired with $p$.
For Porto and Tokyo, a path may not have any candidate paths due to the limited number of paths.
Thus we keep randomly selecting two paths and put them into $D_{pp}$ if their WJ or SWJ is larger than 0.5, and we finally collect 1 million path pairs.
Note that in $D_{pp}$, we remove all the departure time information from each path before PRL, so that the evaluation of the similarity prediction focuses on how well the road segments static information and correlations are captured.  
After the PRL process with $D_{pp}$, we concatenate the representations of two paired paths and train a linear regression (LR) model to predict the WJ or SWJ similarity between the two paths with the concatenated representation.
Specifically, for TraveL, we concatenate $\mu_p$ and $\sigma_p$ to form a vector representation of $p$, which is then fed to the LR model to facilitate training.
Finally, we exploit Mean Absolute Error (MAE) and Mean Absolute Percentage Error (MAPE), between the predicted and the ground truth similarity scores of the path pairs for testing, as the evaluation metrics.
In addition, for Syn-Porto, we evaluate the path representations by ranking all candidate paths, for a given path, in terms of the similarity, and use Kendall rank correlation coefficient (denoted by $\tau$) and Spearman’s rank correlation coefficient (denoted by $\rho$) to measure the consistency between the ranking derived by the predicted similarity scores and the ranking
derived by the ground truth similarity scores. Higher $\tau$ and $\rho$ indicate higher accuracy.

\begin{table}[t]
\caption{Path Similarity Prediction on $D_{mix}$}
\vspace{-0.15in}
\begin{center}
\resizebox{\linewidth}{!}{
\begin{tabular}{|c|c|c|c|c|c|c|c|c|}
\hline
\multicolumn{1}{|c|}{ }&\multicolumn{4}{c|}{Weighted Jaccard}&\multicolumn{4}{c|}{Speed-relevant  Weighted Jaccard}\\
\hline
\multicolumn{1}{|c|}{Model }&\multicolumn{1}{c|}{MAE}&\multicolumn{1}{c|}{MAPE}&\multicolumn{1}{c|}{$\tau$}&\multicolumn{1}{c|}{$\rho$}&\multicolumn{1}{c|}{MAE}&\multicolumn{1}{c|}{MAPE}&\multicolumn{1}{c|}{$\tau$}&\multicolumn{1}{c|}{$\rho$}\\
\cline{1-9}
Node2Vec&
0.18 & 30.1 & 0.62& 0.66 & 0.25 & 40.4 & 0.52 & 0.58\\
\cline{1-9}
RS&
0.16 & 24.8 & 0.66 & 
0.73 & 0.23 & 36.6& 
0.58 & 0.64 \\
\cline{1-9}
BERT &
0.24 & 38.7 & 0.50& 
0.53& 0.27 & 44.9& 
0.44 & 0.48 \\
\cline{1-9}
InfoGraph &
0.17 & 29.5 & 0.65& 
0.69 & 0.21 & 34.8& 
0.50 & 0.55 \\
\cline{1-9}
PIM &
0.12$^*$ & 17.9$^*$ & 0.73$^*$ & 
0.77$^*$ & 0.16 & 24.1& 
0.64 & 0.69 \\
\cline{1-9}
Trembr &
0.13 & 19.8 & 0.72 & 
0.75 & 0.15$^*$ & 21.3$^*$ & 
0.66$^*$ & 0.70$^*$ \\
\cline{1-9}
TraveL &
\textbf{0.10} & \textbf{14.4} & \textbf{0.76}& 
\textbf{0.79} & \textbf{0.11} & \textbf{15.1} & 
\textbf{0.70} & \textbf{0.74} \\
\hline
\end{tabular}}
\label{tab:syn-path-similarity-prediction}
\end{center}
\end{table}

\noindent \textbf{Evaluation of Models.}
As shown in Table~\ref{tab:syn-path-similarity-prediction}, all PRL models perform worse for SWJ prediction on $D_{mix}$, which validates our intuition that SWJ is more challenging to predict owing to its requirement to capture speed-relevant information.
By comparing Trembr with PIM, we observe that Trembr loses for WJ prediction, while wins for SWJ prediction. The reason may be that Trembr captures the same-type relationship (which, e.g., the road segment is on highway, may indicate the speeds) into road segment embeddings to help SWJ prediction.
Finally, owing to the ability to capture varied traveler behaviors, i.e., the varied travel times (which is relevant to speeds), TraveL performs the best among all PRL models by reducing MAE by 16.7\% (0.10 v.s. 0.12 by PIM) for WJ, and MAE by 26.7\% (0.11 v.s. 0.15 by Trembr) for SWJ.
The results for MAPE, $\tau$ and $\rho$ also suggest similar insights as MAE. 
Besides, we have similar observations from the evaluation results on both Porto and Tokyo (as shown in Table~\ref{tab:real-path-similarity-prediction}), which validates our ideas again.
\begin{table}[!ht]
\vspace{-0.15in}
\caption{Path Similarity Prediction on Real-world Datasets}
\vspace{-0.2in}
\begin{center}
\resizebox{\linewidth}{!}{ 
\begin{tabular}{|c|c|c|c|c|c|c|c|c|}
\hline
\multicolumn{1}{|c|}{-}&\multicolumn{2}{c|}{Porto-WJ}&\multicolumn{2}{c|}{Porto-SWJ}&\multicolumn{2}{c|}{Tokyo-WJ}&\multicolumn{2}{c|}{Tokyo-SWJ}\\
\hline
\multicolumn{1}{|c|}{Model }&\multicolumn{1}{c|}{MAE}&\multicolumn{1}{c|}{MAPE}&\multicolumn{1}{c|}{MAE}&\multicolumn{1}{c|}{MAPE}&\multicolumn{1}{c|}{MAE}&\multicolumn{1}{c|}{MAPE}&\multicolumn{1}{c|}{MAE}&\multicolumn{1}{c|}{MAPE}\\
\cline{1-9}
Node2Vec&
0.19 & 30.1 & 0.21 & 
33.0 & 0.17 & 26.6& 
0.20 & 31.2\\
\cline{1-9}
RS&
0.21 & 32.0 & 0.22 & 
35.4 & 0.15 & 24.8 & 
0.17 & 27.1\\
\cline{1-9}
BERT &
0.23 & 36.2 & 0.27 & 41.9 & 0.22 & 35.3 & 
0.26 & 40.3\\
\cline{1-9}
InfoGraph &
0.22 & 35.7 & 0.25 & 37.7 & 0.20 & 32.1& 
0.35 & 39.0\\
\cline{1-9}
PIM &
0.17 & 28.1 & 0.19 & 
31.8 & 0.14 & 22.3 & 
0.26 & 28.5\\
\cline{1-9}
Trembr &
0.15$^*$ & 24.6$^*$ & 0.18$^*$ & 
28.6$^*$ & 0.13$^*$ & 20.4$^*$ & 
0.16$^*$ & 24.9$^*$\\
\cline{1-9}
TraveL &
\textbf{0.13} & \textbf{19.9} & \textbf{0.14} & 
\textbf{21.9} & \textbf{0.12} & \textbf{17.5} & 
\textbf{0.13} & \textbf{20.8}\\
\hline
\end{tabular}
}
\label{tab:real-path-similarity-prediction}
\end{center}
\vspace{-0.3in}
\end{table}

\subsection{Destination Prediction (DP)} 
Here we evaluate all PRL models for the DP task by exploiting a partial path's representation to predict the path destination, i.e., the coordinate of the end of the last road segment in the synthetic dataset or the last coordinate generated by map matching for real-world datasets.
For each dataset, given a path $p$, we generate a partial path starting from $p$'s origin with length as $\lfloor \delta \times |p|\rfloor$, where $\delta$ is the ratio to control the partial path length.
We train all PRL models to obtain the representation for each partial path.
Similar to the evaluation process of PSP, 
we split all partial paths as 90\% and 10\% as the training and testing set respectively, and train two LR models to map the partial path representation to the latitude and longitude of the destination respectively, with the training set.
On the testing set, we calculate the MAE of the geographic distance between the predicted coordinate and the destination in kilometers as the metric for DP. 
Finally, we vary $\delta$ as 0.5, 0.625, 0.75 and 0.875 to evaluate its impacts on the model performance.

\begin{figure}[t]
\centering
\includegraphics[width=3.36in]{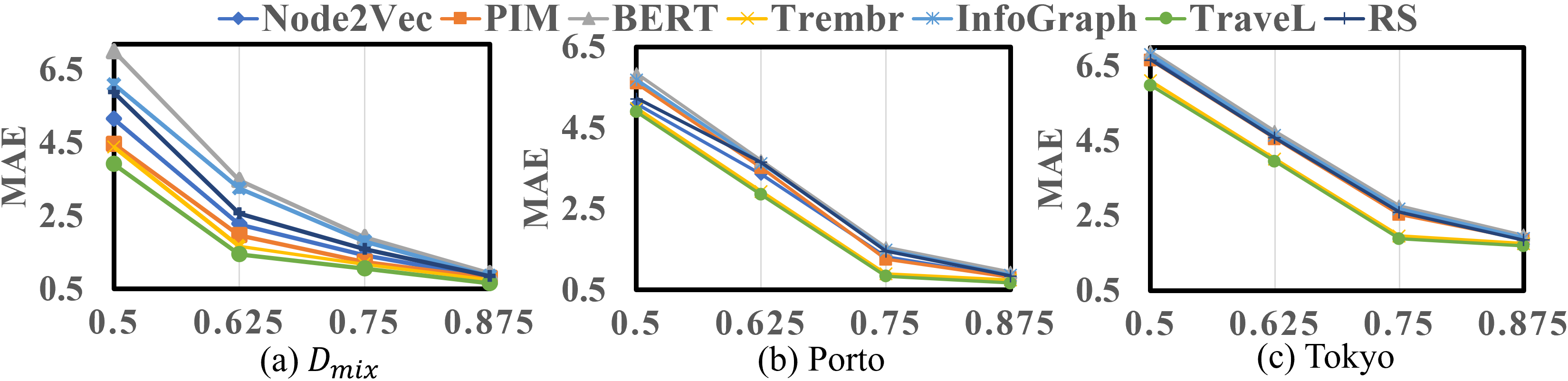}
\vspace{-0.3in}
\caption{Evaluation of Destination Prediction} 
\label{fig:destination-prediction}
\end{figure}

Figure~\ref{fig:destination-prediction} shows the results of DP on all three datasets. We have two main observations: i) as $\delta$ increases, all models achieve a significantly lower MAE, which suggests the decrease of the difficulty of the task; and ii) Although DP is intuitively the least relevant, among the three targeted tasks, to the travel time information, TraveL (i.e., the green line) still outperforms other baseline models consistently on all the three datasets, by reducing MAE (when $\delta=0.75$) from 3.97\% (i.e., 1.69 v.s. 1.76 by Trembr on Tokyo) to 10.6\% (0.67 v.s. 0.75 by Trembr on Porto), which indicates the importance of capturing the regional correlations and the varied traveler behaviors for PRL.

\subsection{Ablation Study}

In this section, we perform an ablation study to assess the impacts of various components in TraveL on its performance. We consider the following variants of TraveL by removing some components from TraveL framework.
i) \texttt{Complete}: the complete version of TraveL;
ii), \texttt{NoViews}: removing all the multi-view attention layers from Path Encoder; iii) \texttt{NoHW}, removing Highway-view attention; iv) \texttt{NoLane}: removing Lane-view attention; v) \texttt{NoHop}: removing Hop-view attention; 
vi) \texttt{NoDistr}: replacing the distribution representation by a vector representation;
and vii) \texttt{OnlyPath}: with OP-Seq Generator predicting the next road segments without generating travel times.

Table~\ref{tab:ablation-study} shows the results of the ablation study on the three datasets for all three tasks. We observe that \texttt{Complete} outperforms all variants on the PSP task.
Between \texttt{OnlyPath} and \texttt{NoViews}, we observe the former has more performance deterioration against \texttt{Complete} for PSP than the
latter, indicating that capturing varied traveler behaviors is more important than capturing the regional correlation for PSP. Among \texttt{NoViews}, \texttt{NoHW}, \texttt{NoLane} and \texttt{NoHop}, we observe that \texttt{NoViews} always performs the worst, which indicates that all these three views benefit PRL in a complementary way.
Besides, we observe that \texttt{NoHop} performs worse than \texttt{NoHW} and \texttt{NoLane} for DP, while \texttt{NoHW} and \texttt{NoLane} perform worse than \texttt{NoHop} for PSP, which suggests that different views may have distinguished benefits for different tasks. Finally, we observe that \texttt{NoDistr} performs the worst (for TTDE and DP) and the second worst (for PSP) among all variants, validating our idea that a distributional representation has larger capacity than a vector, which is essential to help capture the varied traveler behaviors for PRL.

\begin{table}[t]
\caption{Ablation Study of TraveL}
\vspace{-0.15in}
\begin{center}
\resizebox{\linewidth}{!}{ 
\begin{tabular}{|c|c|c|c|c|c|c|c|c|c|}
\hline
\multicolumn{1}{|c|}{ }&\multicolumn{3}{c|}{TTDE (MKS)}&\multicolumn{3}{c|}{PSP (MAE) }&\multicolumn{3}{c|}{DP (MAE)}\\
\hline
\multicolumn{1}{|c|}{Model }&\multicolumn{1}{c|}{$D_{mix}$}&\multicolumn{1}{c|}{Porto}&\multicolumn{1}{c|}{Tokyo}&\multicolumn{1}{c|}{$D_{mix}$}&\multicolumn{1}{c|}{Porto}&\multicolumn{1}{c|}{Tokyo}&\multicolumn{1}{c|}{$D_{mix}$}&\multicolumn{1}{c|}{Porto}&\multicolumn{1}{c|}{Tokyo}\\
\cline{1-10}
\texttt{Complete} &
0.29 & 0.32 & 0.28& 
0.11 & 0.14 & 0.13& 
1.06 & 0.84 & 1.88\\
\cline{1-10}
\texttt{NoViews} &
0.31 & 0.33 & 0.30 & 
0.14 & 0.17 & 0.14& 
1.15 & 0.88 & 1.94\\
\cline{1-10}
\texttt{NoHW} &
0.30 & 0.32 &0.29& 
0.14 & 0.15 & 0.13& 
1.10 & 0.86 & 1.92\\
\cline{1-10}
\texttt{NoLane} &
0.31 & 0.33 & 0.30& 
0.14 & 0.15 & 0.14& 
1.08 & 0.84 & 1.89\\
\cline{1-10}
\texttt{NoHop} &
0.30 & 0.33 & 0.30 & 
0.12 & 0.14 & 0.13& 
1.12 & 0.85 & 1.90\\
\cline{1-10}
\texttt{NoDistr} &
0.32 & 0.34 & 0.33& 
0.14 & 0.17 & 0.16& 
1.16 & 0.87 & 1.93\\
\cline{1-10}
\texttt{OnlyPath} &
0.31 & 0.34 & 0.31 & 
0.13 & 0.18 & 0.17& 
1.09 & 0.89 & 1.91\\
\hline
\end{tabular}
}
\label{tab:ablation-study}
\end{center}
\end{table}

\subsection{Details of Kolmogorov–Smirnov Test}
Kolmogorov–Smirnov Test is a classic test of the equality of two continuous one-dimensional probability distributions based on two sets of values sampled from the two distributions respectively. 
For example, given the generated travel time set ($S^\prime_p$) and the historical travel time set ($S_p$), K-S test transforms each set to an empirical distribution function, i.e., 
$F_n(x) = \frac{1}{n}\sum^n_{i=1}{\textbf{1}_{[-\infty, x]}(X_i)}$,
where $X_i$ is the $i$-th smallest value in $S^\prime_p$, $\textbf{1}_{[-\infty, x]}(X_i)$ is the indicator function, equal to 1 if $X_i \leq x$ and equal to 0 otherwise. And similarly, we obtain $F_m(x)$ from $S_p$.
Then K-S test defines the K-S distance between the two sample sets as $KS(S^\prime_p, S_p) = {\sup}_x|F_n(x)-F_m(x)|$,
which is then exploited to judge whether the two sample sets are from a same distribution.

\subsection{Hyper-parameter Settings}
In this section, we detail the optimal hyper-parameter setting of TraveL: i) the number of multi-view regional attention layers (denoted as $L$) in Multi-view Path Transformer as 6; ii) the number of hops in Hop-view as 3; iii) the dimensionality of $\mu_p$ and $\sigma_p$ as 128; and vi) the number of sample points from the distributional representation as 100. 
For Syn-porto, we have $\lambda_1$ as 0.3, $\lambda_2$ as 0.0 (since synthetic data does not face the data sparsity issue), $\lambda_3$ as 0.7, $\lambda_4$ as 0.1 and $\lambda_5$ as 0.05 to weight the loss. For Porto and Tokyo, we have $\lambda_1$ as 0.1, $\lambda_2$ as 0.3, $\lambda_3$ as 0.6, $\lambda_4$ as 0.1 and $\lambda_5$ as 0.05.
We conduct parameter sensitivity tests to evaluate the impacts of these parameters.
We observe that increasing $L$ from 6 to 8 leads to a worse performance of TraveL on all three tasks on Porto and Tokyo, but better performances on $D_{mix}$. The reason may be that the larger amount of synthetic data (than those of the real-world datasets) supports training a deeper neural network. Other details are not shown due to the space limit.

\section{Conclusion}\label{Conclusion}

We propose the TraveL framework for path representation learning in the road network, where the idea of distributional representation is explored, together with a sampling based On-path Sequence Generator, to capture varied traveler behaviors on the path, and a multi-view regional attention is developed to capture various correlations within regions of road segments. We explore the idea of K-S test to facilitate model training and evaluation.
Empirically, we demonstrate the superiority of TraveL to the state of the arts. 
As for our next step, we plan to explore distributional representations in other forms to capture other types of traveler behaviors for PRL.


\bibliographystyle{ACM-Reference-Format}
\bibliography{citation}


\end{document}